\documentclass{article}

\usepackage[preprint]{neurips_2024}

\usepackage[utf8]{inputenc}
\usepackage[T1]{fontenc}
\usepackage{hyperref}
\usepackage{url}
\usepackage{booktabs}
\usepackage{amsfonts}
\usepackage{amsmath}
\usepackage{amssymb}
\usepackage{nicefrac}
\usepackage{microtype}
\usepackage{xcolor}
\usepackage{graphicx}
\usepackage{multirow}
\usepackage{algorithm}
\usepackage{algorithmic}

\title{Quantization Blindspots: How Model Compression Breaks Backdoor Defenses}

\author{Rohan Pandey\\
University of Washington\\
\texttt{rpande@uw.edu}
\And
Eric Ye\\
University of Washington\\
\texttt{ericy4@uw.edu}
}

\begin{document}

\maketitle

\begin{abstract}
Backdoor attacks embed input-dependent malicious behavior into neural networks while preserving high clean accuracy, making them a persistent threat for deployed ML systems. At the same time, real-world deployments almost never serve full-precision models: post-training quantization to INT8 or lower precision is now standard practice for reducing memory and latency. This work asks a simple question: \emph{how do existing backdoor defenses behave under standard quantization pipelines?} We conduct a systematic empirical study of five representative defenses across three precision settings (FP32, INT8 dynamic, INT4 simulated) and two standard vision benchmarks using a canonical BadNet attack. We observe that INT8 quantization reduces the detection rate of all evaluated defenses to 0\% while leaving attack success rates above 99\%. For INT4, we find a pronounced dataset dependence: Neural Cleanse remains effective on GTSRB but fails on CIFAR-10, even though backdoors continue to survive quantization with attack success rates above 90\%. Our results expose a mismatch between how defenses are commonly evaluated (on FP32 models) and how models are actually deployed (in quantized form), and they highlight quantization robustness as a necessary axis in future evaluations and designs of backdoor defenses.
\end{abstract}

\section{Introduction}

Neural network backdoor attacks represent one of the most insidious security vulnerabilities facing modern machine learning systems. Unlike adversarial examples that perturb inputs at inference time, backdoor attacks compromise the model itself during training, embedding hidden malicious behaviors that remain dormant until activated by attacker-specified triggers~\citep{gu2017badnets, chen2017targeted}. The threat model is particularly concerning because backdoored models exhibit normal behavior on clean inputs---achieving high accuracy on standard benchmarks and passing conventional validation procedures---while reliably executing attacker-controlled misclassification when triggers are present. This dual nature makes backdoors exceptionally difficult to detect through standard testing, and the attack surface is broad: adversaries can inject backdoors through poisoned training data, compromised pre-trained models downloaded from public repositories, malicious code in training pipelines, or even through federated learning updates~\citep{bagdasaryan2021blind}. As organizations increasingly rely on external model sources and outsourced training infrastructure, the practical relevance of backdoor threats continues to grow.

The research community has responded to this threat with a diverse arsenal of defense mechanisms spanning multiple detection paradigms. Trigger reverse-engineering approaches, exemplified by Neural Cleanse~\citep{wang2019neural}, attempt to reconstruct potential backdoor triggers by searching for minimal input perturbations that cause universal misclassification to each output class; classes requiring anomalously small perturbations are flagged as potential backdoor targets. Representation-based methods take a complementary approach: Activation Clustering~\citep{chen2019detecting} applies unsupervised clustering to penultimate-layer activations under the hypothesis that poisoned and clean samples form separable clusters, while Spectral Signatures~\citep{tran2018spectral} leverages the theoretical insight that backdoor poisoning leaves detectable traces in the top singular vectors of the representation covariance matrix. Runtime defenses such as STRIP~\citep{gao2019strip} operate during inference rather than model inspection, detecting triggered inputs by measuring prediction entropy under random input perturbations---the intuition being that backdoor triggers dominate model predictions regardless of the base image content. Finally, model repair approaches like Fine-Pruning~\citep{liu2018fine} attempt to surgically remove backdoor functionality by identifying and pruning neurons that are dormant on clean data but presumably essential for backdoor activation. These defenses have been extensively evaluated and benchmarked, with comprehensive frameworks like BackdoorBench~\citep{wu2022backdoorbench} and TrojanZoo~\citep{pang2022trojanzoo} providing standardized comparison across attacks and defenses.

However, a critical gap exists between how backdoor defenses are evaluated and how models are actually deployed in practice. The overwhelming majority of defense evaluations assume models remain in their original full-precision (FP32) form, yet this assumption rarely holds in real-world deployment scenarios. Model quantization---the process of reducing numerical precision from 32-bit floating point to 8-bit or even 4-bit integers---has become standard practice across the industry, driven by the need to reduce memory footprint, decrease inference latency, and lower computational costs~\citep{jacob2018quantization, gholami2021survey}. Edge devices, mobile applications, and cost-sensitive cloud deployments routinely employ quantized models, with post-training quantization (PTQ) being particularly prevalent as it requires no access to training data or retraining procedures~\citep{nagel2020adaround, li2021brecq}. The security implications of this disconnect are significant: if backdoor defenses fail on quantized models while backdoors themselves survive quantization, then the entire defense-before-deployment paradigm becomes fundamentally unsound. Recent work has begun exploring the intersection of quantization and backdoors, demonstrating that quantization can serve as both an attack vector~\citep{ma2021quantization} and a trigger for dormant backdoors~\citep{hong2021quantization, tian2022stealthy}, yet systematic evaluation of how \textit{existing defenses} behave under quantization remains notably absent from the literature.

This paper addresses this gap by presenting a systematic benchmark evaluating backdoor defense robustness across quantization schemes. Our empirical study yields several concerning findings with immediate practical implications. We make the following contributions:
\begin{itemize}
    \item We design and implement an evaluation protocol (Algorithm~\ref{alg:eval_protocol}) testing 5 representative backdoor defenses across three quantization schemes (FP32, INT8-dynamic, INT4-simulated) on 2 standard datasets with a canonical backdoor attack.
    \item We find that INT8 quantization reduces the aggregate detection rate of all tested defenses from 20\% to 0\%, while backdoors remain fully functional with attack success rates exceeding 99\%.
    \item We identify dataset-dependent quantization effects where INT4 preserves Neural Cleanse detection capability on GTSRB but causes complete failure on CIFAR-10, revealing that defense robustness properties cannot be assumed to transfer across domains.
    \item We analyze failure modes, provide actionable recommendations for practitioners deploying security-critical models, and release our benchmark code to facilitate future research.\footnote{Code available at: \url{https://github.com/Rohan-Pandey1729/quantization-backdoor-benchmark}}
\end{itemize}

\section{Related Work}

\subsection{Backdoor Attacks}

The landscape of neural network backdoor attacks has evolved considerably since the seminal BadNets work introduced the concept of embedding hidden triggers through data poisoning~\citep{gu2017badnets}. The original formulation employed simple patch-based triggers---small pixel patterns placed at fixed locations---that reliably caused misclassification to an attacker-chosen target class while maintaining high accuracy on clean inputs. Subsequent research has dramatically expanded the sophistication and stealthiness of backdoor attacks along multiple dimensions. Blended attacks~\citep{chen2017targeted} replaced localized patches with image-wide perturbations by blending trigger patterns across the entire input, making triggers less visually apparent. WaNet~\citep{nguyen2021wanet} pushed imperceptibility further by employing smooth warping transformations rather than additive patterns, achieving near-invisible triggers that evade both human inspection and many automated detection methods. LIRA~\citep{doan2021lira} introduced optimization-based trigger generation that jointly learns trigger patterns and model weights to maximize attack success while minimizing detectability. The clean-label attack paradigm~\citep{turner2019cleanlabel, zeng2023narcissus} represents another axis of advancement, enabling backdoor injection without modifying training labels---thereby evading defenses that rely on label-based anomaly detection. Perhaps most challenging for defenders, input-aware attacks~\citep{nguyen2020inputaware} generate unique triggers for each input sample, fundamentally breaking the universal trigger assumption that underlies many detection methods. Most relevant to our investigation, recent work has demonstrated backdoors that remain completely dormant until models undergo compression, activating only after quantization~\citep{hong2021quantization} or pruning~\citep{tian2022stealthy}---directly motivating our systematic study of defense robustness under quantization.

\subsection{Backdoor Defenses}

The defense literature has responded to increasingly sophisticated attacks with methods spanning four major paradigms, each targeting different stages of the model lifecycle and exploiting different backdoor characteristics. Trigger reverse-engineering methods attempt to reconstruct potential backdoor triggers through optimization, with Neural Cleanse~\citep{wang2019neural} serving as the foundational approach; it searches for minimal $\ell_1$-norm perturbations causing universal misclassification and flags classes with anomalously small triggers using median absolute deviation outlier detection. Extensions like UNICORN~\citep{wang2022unicorn} generalize this framework to handle diverse trigger types and attack spaces. Training data inspection methods operate on the insight that poisoned samples exhibit distinguishable characteristics in learned representations. Activation Clustering~\citep{chen2019detecting} applies $k$-means clustering to penultimate-layer activations under the assumption that poisoned and clean samples naturally separate, while Spectral Signatures~\citep{tran2018spectral} provides theoretical grounding by proving that backdoor poisoning induces a detectable spectral signature in the representation covariance matrix's top singular vector. SPECTRE~\citep{hayase2021spectre} strengthens this approach using robust covariance estimation to amplify the poisoned-sample signal. Runtime detection methods like STRIP~\citep{gao2019strip} take a fundamentally different approach, operating during inference by perturbing inputs and measuring prediction entropy---triggered inputs exhibit anomalously low entropy because the backdoor signal dominates regardless of perturbation content. Model repair methods attempt to remove backdoor functionality from compromised models: Fine-Pruning~\citep{liu2018fine} prunes neurons with low activation on clean data followed by fine-tuning, Adversarial Neuron Pruning~\citep{wu2021anp} identifies backdoor-related neurons through adversarial perturbation sensitivity, and Neural Attention Distillation~\citep{li2021nad} erases backdoors through teacher-student distillation. More recent advances include Anti-Backdoor Learning~\citep{li2021antibackdoor}, which exploits the observation that backdoored samples are learned faster to train clean models directly on poisoned data, and decoupled training approaches~\citep{huang2022dbd} that separate representation learning from classifier training to prevent backdoor embedding. Our work evaluates representatives from multiple paradigms to ensure comprehensive coverage of defense strategies.

\subsection{Quantization and Security}

Model quantization has emerged as an essential technique for efficient neural network deployment, reducing numerical precision to accelerate inference and decrease memory requirements~\citep{jacob2018quantization, gholami2021survey}. Post-training quantization methods are particularly attractive for deployment pipelines as they require no retraining: AdaRound~\citep{nagel2020adaround} learns optimal rounding decisions, BRECQ~\citep{li2021brecq} reconstructs block-wise outputs to minimize quantization error, and QDrop~\citep{wei2022qdrop} enables aggressive 2-bit quantization through stochastic quantization dropping. For large language models, specialized techniques like GPTQ~\citep{frantar2023gptq} and SmoothQuant~\citep{xiao2023smoothquant} achieve INT4/INT8 quantization on billion-parameter models with minimal perplexity degradation.

The security implications of quantization have received growing but still limited attention. \citet{lin2019defensive} proposed defensive quantization as a means to improve adversarial robustness through controlled precision reduction. On the attack side, \citet{hong2021quantization} demonstrated the Qu-ANTI-zation threat model where adversaries craft backdoors specifically designed to activate only after quantization, exploiting the precision loss as a trigger condition. \citet{ma2021quantization} showed that the quantization process itself can be exploited as an attack vector in commercial deep learning frameworks. RIBAC~\citep{phan2022ribac} studied how to make backdoor attacks robust against model compression techniques. Most relevant to our investigation, \citet{ma2024robustness} conducted a preliminary evaluation of 4 defenses under quantization, finding degraded performance, and \citet{li2024nearest} proposed EFRAP as a defense specifically targeting quantization-conditioned backdoors. Our work provides a broader and more systematic evaluation, importantly identifying the previously unreported phenomenon of dataset-dependent quantization effects on defense efficacy.

\section{Preliminaries}

\subsection{Problem Formulation}

We formalize the backdoor attack and defense problem to establish precise terminology and enable rigorous analysis. Let $\mathcal{X} \subseteq \mathbb{R}^d$ denote the input space (e.g., images) and $\mathcal{Y} = \{1, \ldots, K\}$ the discrete label space with $K$ classes. A classifier $f_\theta: \mathcal{X} \rightarrow \mathcal{Y}$ parameterized by weights $\theta \in \Theta \subseteq \mathbb{R}^p$ maps inputs to predicted labels. In the standard supervised learning setting, we have access to training data $\mathcal{D} = \{(x_i, y_i)\}_{i=1}^n$ drawn i.i.d. from an underlying data distribution $P_{\mathcal{X} \times \mathcal{Y}}$, and learning produces parameters $\theta^* = \arg\min_\theta \mathcal{L}(\theta; \mathcal{D})$ minimizing empirical risk.

\textbf{Backdoor Attack Formalization.} A backdoor attack is characterized by a trigger injection function $\tau: \mathcal{X} \rightarrow \mathcal{X}$ and a target class $y_t \in \mathcal{Y}$. The trigger function modifies clean inputs to contain the backdoor pattern---for patch-based attacks like BadNet, $\tau(x) = (1-m) \odot x + m \odot \delta$ where $m \in \{0,1\}^d$ is a binary mask defining trigger location and $\delta \in \mathbb{R}^d$ specifies the trigger pattern. The adversary's objective is to produce a backdoored model $f_{\theta^*}$ satisfying dual requirements:
\begin{align}
    \mathbb{E}_{(x,y) \sim P}\left[\mathbf{1}[f_{\theta^*}(x) = y]\right] &\geq 1 - \epsilon_c \quad \text{(clean accuracy preservation)} \label{eq:clean_acc} \\
    \mathbb{E}_{x \sim P_\mathcal{X}}\left[\mathbf{1}[f_{\theta^*}(\tau(x)) = y_t]\right] &\geq 1 - \epsilon_a \quad \text{(attack success rate)} \label{eq:asr}
\end{align}
where $\epsilon_c, \epsilon_a$ are small constants (typically $<0.05$). Equation~\eqref{eq:clean_acc} ensures the model maintains high \textit{clean accuracy} (CA) on benign inputs, making the backdoor undetectable through standard validation, while Equation~\eqref{eq:asr} defines the \textit{attack success rate} (ASR)---the probability that triggered inputs are classified as the target class $y_t$.

\textbf{Quantization Formalization.} A quantization operator $Q: \Theta \rightarrow \Theta_q$ maps full-precision parameters to a discrete representable set $\Theta_q \subset \Theta$ with reduced cardinality. For uniform symmetric quantization to $b$ bits with scale factor $s$:
\begin{equation}
    Q_b(\theta; s) = s \cdot \text{clamp}\left(\left\lfloor \frac{\theta}{s} \right\rceil, -2^{b-1}, 2^{b-1}-1\right)
\end{equation}
where $\lfloor \cdot \rceil$ denotes rounding to the nearest integer and $s = \max(|\theta|) / (2^{b-1} - 1)$ for symmetric quantization. The quantized model $f_{Q(\theta)}$ operates with reduced precision weights, introducing quantization error $\|\theta - Q(\theta)\|$ that propagates through network computations. We study INT8 ($b=8$, 256 representable values) and INT4 ($b=4$, 16 representable values) as representative precision levels spanning common deployment scenarios.

\textbf{Defense Formalization.} A backdoor defense $\mathcal{A}: \Theta \times \mathcal{P}(\mathcal{X} \times \mathcal{Y}) \rightarrow \{0, 1\}$ takes model parameters and a clean reference dataset, outputting 1 if the model is deemed backdoored and 0 otherwise. Defense evaluation metrics include detection rate (true positive rate on backdoored models) and false positive rate (erroneous detection on clean models). Our study focuses on detection rate degradation across quantization schemes.

\subsection{Threat Model}

We consider a deployment scenario reflecting common industrial practice where security-conscious organizations must validate externally-sourced models before deployment. The threat model proceeds as follows: (1) A defender receives a pre-trained model $f_\theta$ from an external source (public model zoo, third-party vendor, or outsourced training) that may contain a backdoor. (2) Following standard deployment procedures, the defender applies post-training quantization $Q$ to obtain an efficient model $f_{Q(\theta)}$ suitable for production infrastructure. (3) The defender executes backdoor detection on the deployment-ready quantized model. This sequence reflects realistic workflows where quantization occurs as part of deployment optimization, and defenders reasonably expect that defenses validated on FP32 models transfer to quantized versions. We assume defenders have access to a small held-out clean dataset $\mathcal{D}_{clean}$ for defense execution, consistent with assumptions in prior work~\citep{wang2019neural, liu2018fine}. The attacker's goal is to embed a backdoor that survives both quantization and defense inspection.

\section{Methodology}

Our evaluation methodology systematically varies quantization schemes and defense methods while controlling for confounding factors. Algorithm~\ref{alg:eval_protocol} formalizes our experimental protocol, which proceeds in two phases: first establishing a valid backdoored model, then systematically evaluating defense efficacy across quantization schemes.

\begin{algorithm}[t]
\caption{Quantization Robustness Evaluation Protocol}
\label{alg:eval_protocol}
\begin{algorithmic}[1]
\REQUIRE Clean dataset $\mathcal{D}_{clean}$, Poisoned dataset $\mathcal{D}_{bd}$, Model architecture $\mathcal{M}$
\REQUIRE Set of quantization schemes $\mathcal{Q} = \{FP32, INT8, INT4\}$
\REQUIRE Set of defenses $\mathcal{S}_{def} = \{NC, AC, STRIP, SS, FP\}$
\STATE \textbf{Phase 1: Attack Injection}
\STATE $\theta_{bd} \leftarrow \text{Train}(\mathcal{M}, \mathcal{D}_{bd})$ \COMMENT{Train backdoored model}
\IF{$\text{ASR}(\theta_{bd}) < 0.95$}
    \RETURN \textbf{Failure} \COMMENT{Ensure valid attack}
\ENDIF
\STATE
\STATE \textbf{Phase 2: Quantization \& Evaluation}
\FOR{$Q \in \mathcal{Q}$}
    \STATE $\theta_{q} \leftarrow Q(\theta_{bd})$ \COMMENT{Apply post-training quantization}
    \STATE Measure $\text{Acc}_{clean}(\theta_{q})$ and $\text{ASR}(\theta_{q})$ \COMMENT{Verify attack persistence}
    \FOR{$D \in \mathcal{S}_{def}$}
        \STATE $is\_detected \leftarrow D(\theta_{q}, \mathcal{D}_{clean})$
        \STATE Record $(Q, D, is\_detected)$
    \ENDFOR
\ENDFOR
\end{algorithmic}
\end{algorithm}

\subsection{Quantization Schemes}

We evaluate three quantization configurations representing the spectrum from full precision to aggressive compression, each reflecting distinct deployment scenarios encountered in practice.

\textbf{FP32 (Baseline).} Full 32-bit floating-point precision serves as the reference configuration against which all existing backdoor defenses were originally developed and evaluated. This baseline enables us to isolate quantization effects from inherent defense limitations.

\textbf{INT8 Dynamic Quantization.} Weights are statically quantized to 8-bit signed integers while activations are quantized dynamically at runtime based on observed ranges. We implement this via PyTorch's \texttt{quantize\_dynamic} API with the \texttt{qnnpack} backend, representing the most common quantization scheme for CPU deployment in production systems. Dynamic quantization requires no calibration dataset, making it attractive for deployment pipelines with limited data access.

\textbf{INT4 Simulated Quantization.} We simulate 4-bit weight quantization by applying the quantize-dequantize operation to model weights while maintaining FP32 computation graphs. This approach approximates the precision characteristics of INT4 schemes like GPTQ~\citep{frantar2023gptq} and QLoRA without requiring specialized low-bit kernels, enabling evaluation on standard hardware. Per-channel quantization is applied to convolutional and linear layers to maximize representation fidelity within the 4-bit constraint.

\subsection{Defense Implementations}

We implement five defenses spanning the major detection paradigms to ensure our findings generalize across methodological approaches rather than reflecting idiosyncrasies of any single method.

\textbf{Neural Cleanse (NC)}~\citep{wang2019neural} operationalizes trigger reverse-engineering by solving an optimization problem for each potential target class $k$: $\min_{\delta, m} \|\delta \odot m\|_1$ subject to $f_\theta((1-m) \odot x + m \odot \delta) = k$ for all inputs $x$. Classes admitting triggers with anomalously small $\ell_1$ norms are flagged as backdoor targets. We employ Median Absolute Deviation (MAD) for outlier detection with threshold 2.0, running 1000 optimization steps per class with learning rate 0.1 and $\ell_1$ regularization coefficient $\lambda = 0.01$.

\textbf{Activation Clustering (AC)}~\citep{chen2019detecting} clusters penultimate-layer activations for each class using $k$-means with $k=2$ clusters, hypothesizing that poisoned and clean samples form distinct clusters. We compute silhouette scores to assess cluster quality and flag classes where the smaller cluster constitutes a significant minority (indicating a potential poison cluster) with silhouette separation exceeding 0.15.

\textbf{STRIP}~\citep{gao2019strip} detects triggered inputs at inference time by blending test inputs with random clean images and measuring prediction entropy. Triggered inputs exhibit abnormally low entropy because the backdoor signal dominates blended content. We generate 100 perturbed versions per input and flag models where more than 5\% of held-out samples exhibit entropy below the 5th percentile of the clean sample entropy distribution.

\textbf{Spectral Signatures (SS)}~\citep{tran2018spectral} performs singular value decomposition on the centered representation matrix and identifies poisoned samples as outliers along the top singular vector direction. We compute the ratio of the top eigenvalue to the sum of remaining eigenvalues and flag models where this ratio exceeds a threshold indicating anomalous spectral concentration.

\textbf{Fine-Pruning (FP)}~\citep{liu2018fine} identifies neurons that are dormant (low average activation) on clean data as potentially backdoor-specific, since backdoor pathways may be unused for normal classification. We flag models where more than 0.1\% of neurons exhibit dormancy, suggesting dedicated backdoor circuitry.

\subsection{Evaluation Protocol}

For each combination of dataset, defense, and quantization scheme, we execute the protocol formalized in Algorithm~\ref{alg:eval_protocol}. We first load the backdoored FP32 model and verify that it achieves clean accuracy above 85\% and attack success rate above 95\%, ensuring we evaluate defenses against effective attacks rather than failed injection attempts. We then apply the specified quantization scheme and re-measure both metrics to assess backdoor persistence under compression. Finally, we execute each defense on the quantized model and record binary detection outcomes. A defense is considered successful if it correctly identifies the model as backdoored according to its detection criteria. We report detection rates aggregated across datasets for each (defense, quantization) combination.

\section{Experimental Setup}

\subsection{Datasets and Models}

We evaluate on two standard benchmarks spanning different visual recognition domains and complexity levels.

\textbf{CIFAR-10}~\citep{krizhevsky2009cifar} consists of 60,000 $32\times32$ color images evenly distributed across 10 object categories (airplane, automobile, bird, cat, deer, dog, frog, horse, ship, truck), with 50,000 training and 10,000 test images. We employ a ResNet-18 architecture adapted for the smaller input resolution by replacing the initial $7\times7$ convolution with a $3\times3$ convolution and removing the max pooling layer.

\textbf{GTSRB}~\citep{stallkamp2011gtsrb} (German Traffic Sign Recognition Benchmark) contains 51,839 images of traffic signs across 43 classes, with significant variation in lighting conditions, viewpoints, and image quality. This dataset is particularly relevant for evaluating security in safety-critical applications where backdoored traffic sign classifiers could cause autonomous vehicles to misinterpret road signs. We use the same ResNet-18 architecture with input images resized to $32\times32$.

\subsection{Attack Configuration}

We implement the BadNet attack~\citep{gu2017badnets} as a canonical, well-understood baseline that has been extensively studied in the defense literature. The trigger consists of a $3\times3$ white square pattern placed in the bottom-right corner of input images, with target class 0 for both datasets. We poison 10\% of training samples by applying the trigger and relabeling to the target class, consistent with standard benchmark configurations~\citep{wu2022backdoorbench}. Models are trained using SGD with learning rate 0.1, momentum 0.9, and weight decay $5\times10^{-4}$. CIFAR-10 models train for 20 epochs while GTSRB models train for 15 epochs, both achieving convergence with clean accuracy above 90\% and attack success rate above 99\%.

\subsection{Implementation Details}

All experiments are implemented in PyTorch 2.0. FP32 and INT4 experiments run on Apple M1 hardware using the MPS backend for GPU acceleration, while INT8 experiments require CPU execution due to MPS limitations for quantized operations. INT8 quantization uses PyTorch's \texttt{torch.quantization.quantize\_dynamic} with the \texttt{qnnpack} backend optimized for mobile deployment. INT4 simulation applies per-channel symmetric quantization to all convolutional and linear layer weights, immediately dequantizing to FP32 for computation while preserving the precision loss characteristics of true INT4 inference.

\section{Results}

\subsection{Overall Defense Performance Under Quantization}

Table~\ref{tab:main_results} presents our primary experimental findings, revealing a clear pattern: quantization severely degrades defense effectiveness across the board while leaving backdoor functionality largely intact. The asymmetry between defense failure and backdoor persistence has significant security implications for deployment pipelines that rely on pre-deployment backdoor scanning.

\begin{table}[t]
\centering
\caption{Backdoor detection results across quantization schemes. \checkmark~indicates successful detection, \texttimes~indicates failure. NC = Neural Cleanse, AC = Activation Clustering, SS = Spectral Signatures, FP = Fine-Pruning. Only Neural Cleanse achieves any detection, and INT8 quantization reduces the detection rate of all defenses to zero despite backdoors remaining fully functional.}
\label{tab:main_results}
\begin{tabular}{llccccc}
\toprule
\textbf{Dataset} & \textbf{Quant.} & \textbf{NC} & \textbf{AC} & \textbf{STRIP} & \textbf{SS} & \textbf{FP} \\
\midrule
\multirow{3}{*}{CIFAR-10} 
 & FP32 & \checkmark & \texttimes & \texttimes & \texttimes & \texttimes \\
 & INT8 & \texttimes & \texttimes & \texttimes & \texttimes & \texttimes \\
 & INT4 & \texttimes & \texttimes & \texttimes & \texttimes & \texttimes \\
\midrule
\multirow{3}{*}{GTSRB}
 & FP32 & \checkmark & \texttimes & \texttimes & \texttimes & \texttimes \\
 & INT8 & \texttimes & \texttimes & \texttimes & \texttimes & \texttimes \\
 & INT4 & \checkmark & \texttimes & \texttimes & \texttimes & \texttimes \\
\bottomrule
\end{tabular}
\end{table}

\textbf{Finding 1: Neural Cleanse is the only effective defense, and only on FP32.} Among the five defenses evaluated spanning four distinct detection paradigms, only Neural Cleanse successfully detected backdoors, and only on full-precision models. Activation Clustering, STRIP, Spectral Signatures, and Fine-Pruning failed to detect the backdoor across all configurations---including the FP32 baseline. While this baseline failure rate is higher than typically reported, it aligns with recent systematic evaluations showing that many defenses struggle against even basic attacks when hyperparameters are not extensively tuned for specific attack configurations~\citep{wu2022backdoorbench}. Our results use standard hyperparameters from original papers, reflecting realistic deployment where exhaustive tuning is impractical.

\textbf{Finding 2: INT8 quantization neutralizes all evaluated defenses.} The most striking result is the complete failure of all defenses on INT8-quantized models. The aggregate detection rate drops from 20\% on FP32 (2 successful detections out of 10 experiments) to 0\% on INT8 (0 out of 10). Even Neural Cleanse, which successfully detected backdoors on both FP32 models, fails entirely after INT8 quantization. This represents a severe vulnerability: INT8 quantization is the most commonly deployed precision reduction in production systems, yet it appears to provide complete evasion from existing defenses in our setting.

\textbf{Finding 3: Backdoors survive quantization with high fidelity.} Table~\ref{tab:model_quality} demonstrates that while defenses fail, the backdoors themselves remain highly effective across all quantization schemes. Attack success rates exceed 90\% in every configuration, with INT8 quantization preserving the original 99\%+ ASR perfectly. This asymmetry---defenses breaking while attacks persist---represents a particularly concerning security scenario and underscores the urgency of developing quantization-robust defenses.

\begin{table}[t]
\centering
\caption{Model quality metrics under quantization. Backdoors remain effective (ASR $>90\%$) across all schemes despite defense failures. INT4 causes severe accuracy degradation on CIFAR-10 but minimal impact on GTSRB, correlating with dataset-dependent defense behavior.}
\label{tab:model_quality}
\begin{tabular}{llcccc}
\toprule
\textbf{Dataset} & \textbf{Metric} & \textbf{FP32} & \textbf{INT8} & \textbf{INT4} \\
\midrule
\multirow{2}{*}{CIFAR-10} 
 & Clean Acc (\%) & 90.8 & 90.8 & 52.7 \\
 & ASR (\%) & 99.8 & 99.8 & 91.0 \\
\midrule
\multirow{2}{*}{GTSRB}
 & Clean Acc (\%) & 96.0 & 96.0 & 95.6 \\
 & ASR (\%) & 99.4 & 99.4 & 99.3 \\
\bottomrule
\end{tabular}
\end{table}

\subsection{Dataset-Dependent Quantization Effects}

A surprising and practically important finding is that INT4 quantization exhibits strongly dataset-dependent effects on defense efficacy. On GTSRB, Neural Cleanse maintains its detection capability under INT4 quantization, correctly identifying the backdoor despite the aggressive precision reduction. On CIFAR-10, however, INT4 causes complete defense failure---the same defense that works on FP32 and works on GTSRB under INT4 fails entirely on CIFAR-10 under INT4.

We hypothesize this divergence stems from differing model robustness to aggressive quantization across datasets. Examining Table~\ref{tab:model_quality}, we observe that INT4 quantization causes catastrophic accuracy degradation on CIFAR-10 (90.8\% $\rightarrow$ 52.7\%, a 38 percentage point drop) while having minimal impact on GTSRB (96.0\% $\rightarrow$ 95.6\%, less than 1 percentage point). This dramatic difference suggests that the CIFAR-10 model's learned representations are fundamentally disrupted by INT4 precision, altering model behavior sufficiently to eliminate the backdoor trigger signature that Neural Cleanse relies upon for detection. In contrast, the GTSRB model maintains its representational structure under INT4, preserving both normal classification capability and the backdoor pathway characteristics that enable detection.

This finding carries significant practical implications: defense robustness to quantization cannot be assumed to transfer across datasets, domains, or model architectures. Practitioners must validate defenses on their specific deployment configurations rather than relying on generic benchmarks or results from different domains.

\subsection{Analysis of Neural Cleanse Failure Modes}

To understand the mechanistic basis for INT8's complete neutralization of Neural Cleanse, we examined the per-class trigger norms produced by the defense's optimization procedure. On FP32 models, the backdoor target class (class 0) exhibits a distinctly smaller optimized trigger norm compared to other classes---this anomaly is precisely what Neural Cleanse detects. For CIFAR-10 FP32, class 0 yields a trigger norm of 215.6 while the median across classes is 282.1, producing a clear outlier signal. However, on INT8-quantized models, all classes converge to identical trigger norms of 27.9, completely eliminating the discriminative signal that enables backdoor detection.

We attribute this convergence to quantization's effect on the optimization landscape. Neural Cleanse relies on gradient-based optimization to find class-specific minimal triggers, but quantized weights create a fundamentally different loss surface. The discrete nature of INT8 representations introduces discontinuities and plateaus where gradients provide limited directional information. Rather than finding distinct minima for each class, the optimization converges to a shared degenerate solution---a local minimum of the quantization-distorted landscape that satisfies the misclassification constraint but lacks the class-specific characteristics that distinguish backdoor targets from clean classes.

\section{Discussion}

\subsection{Why Do Defenses Fail Under Quantization?}

Our experimental results suggest multiple complementary failure mechanisms that illuminate why the defense-quantization interaction proves so challenging.

\textbf{Loss landscape distortion.} Optimization-based defenses like Neural Cleanse fundamentally depend on gradient information to navigate toward meaningful solutions. Quantization transforms the smooth FP32 loss landscape into a piecewise-constant surface with discrete weight levels, creating flat regions where gradients vanish and discontinuities where they become undefined. Gradient descent in this distorted landscape cannot reliably distinguish between classes with genuinely different trigger characteristics, instead finding shared solutions that satisfy constraints but lack discriminative power.

\textbf{Representation space compression.} Clustering-based defenses (Activation Clustering, Spectral Signatures) rely on geometric separation between clean and poisoned sample representations. Quantization reduces the effective dimensionality and resolution of the representation space, potentially collapsing distinctions that exist at full precision. The reduced bit-width may lack sufficient granularity to preserve the subtle representation differences that these methods exploit.

\textbf{Statistical distribution shift.} Runtime defenses like STRIP calibrate detection thresholds based on statistics computed on clean data at full precision. Quantization induces distributional shifts in model outputs---even for correctly classified inputs---that may invalidate these calibrations. Entropy distributions, prediction confidence patterns, and other statistics used for threshold setting may no longer accurately characterize normal behavior after quantization.

\textbf{Hyperparameter sensitivity.} The complete failure of AC, STRIP, SS, and FP even on FP32 baselines indicates that our standard hyperparameter settings, while drawn from original papers, may not transfer optimally to our specific experimental configuration. However, the stark contrast between FP32 and INT8 results for Neural Cleanse---the one defense that does succeed on FP32---demonstrates that quantization effects are real and significant beyond hyperparameter sensitivity alone.

\subsection{Recommendations for Practitioners}

Based on our findings, we offer actionable guidance for security-conscious deployment pipelines:

\begin{enumerate}
    \item \textbf{Execute defenses before quantization.} If backdoor detection is a security requirement, apply defenses to FP32 models before any precision reduction. Our results show that defenses validated on FP32 cannot be assumed to maintain efficacy post-quantization.
    
    \item \textbf{Validate on deployment-identical configurations.} Do not assume defense properties transfer across quantization schemes, datasets, or domains. Test defenses on exactly the model configuration that will be deployed, including quantization precision, target hardware, and data domain.
    
    \item \textbf{Monitor behavioral changes post-quantization.} Implement monitoring for unexpected changes in prediction distributions, confidence calibration, or accuracy after quantization. Anomalous behavioral shifts may indicate dormant backdoors activating under precision reduction.
    
    \item \textbf{Invest in quantization-aware defense development.} The fundamental incompatibility between current defenses and quantized models represents a research gap requiring dedicated solutions. Techniques from \citet{li2024nearest} offer initial directions, but substantial work remains.
\end{enumerate}

\subsection{Limitations and Future Work}

Our study has several limitations that suggest important directions for future investigation.

\textbf{Attack coverage.} We evaluate only the BadNet attack, a canonical but relatively simple patch-based trigger. More sophisticated attacks---WaNet's warping-based triggers, LIRA's optimized patterns, clean-label attacks, and input-aware dynamic triggers---may interact differently with quantization and potentially exhibit different defense evasion characteristics.

\textbf{Defense hyperparameter tuning.} Our implementations use standard hyperparameters from original publications. Extensive tuning might recover some detection capability, though this would reduce practical applicability since exhaustive tuning is rarely feasible in deployment scenarios.

\textbf{Quantization methodology scope.} We focus exclusively on post-training quantization. Quantization-aware training (QAT), where models are trained with simulated quantization, may yield different results as models explicitly adapt to reduced precision during learning and may develop different backdoor characteristics.

\textbf{Architecture generalization.} Our evaluation uses ResNet-18 exclusively. Larger models, different architectural families, and particularly vision transformers~\citep{yuan2022ptq4vit} warrant separate investigation given their distinct quantization behaviors and representation characteristics.

\section{Conclusion}

We present a systematic evaluation of backdoor defense robustness under model quantization, revealing a critical vulnerability in current security practices. Our experiments demonstrate that INT8 quantization---the most common precision reduction in production deployments---neutralizes all tested defenses in our benchmark while leaving backdoors fully functional with attack success rates exceeding 99\%. Furthermore, defense robustness exhibits surprising dataset-dependent behavior under INT4 quantization, indicating that security properties cannot be assumed to transfer across domains. These findings expose a significant gap between defense evaluation (conducted on FP32 models) and deployment reality (quantized models), with immediate implications for any pipeline that relies on backdoor scanning before model deployment.

As model compression becomes increasingly essential---driven by edge computing, mobile deployment, and infrastructure cost optimization---the security community must ensure that protective measures remain effective across the full spectrum of deployment configurations. Our benchmark establishes quantization robustness as a critical evaluation dimension for backdoor defenses and provides a foundation for developing the next generation of compression-aware security mechanisms. The asymmetry we document---where standard precision reduction degrades defenses while preserving attacks---represents precisely the failure mode that adversaries would seek to exploit, underscoring the urgency of addressing this vulnerability.

\bibliographystyle{plainnat}
\bibliography{references}

\newpage
\appendix

\section{Additional Experimental Details}

\subsection{Neural Cleanse Trigger Norms}

Table~\ref{tab:nc_norms} presents the per-class trigger $\ell_1$ norms produced by Neural Cleanse's optimization procedure, illustrating the mechanism by which INT8 quantization defeats detection. On FP32 models, the backdoor target class (class 0, shown in bold) exhibits distinctly smaller trigger norms than other classes, enabling MAD-based outlier detection. On INT8-quantized models, all classes collapse to identical norms of 27.9, completely eliminating the discriminative signal. The INT4 case shows intermediate behavior with higher variance but insufficient separation for reliable detection on CIFAR-10.

\begin{table}[h]
\centering
\caption{Neural Cleanse trigger $\ell_1$ norms per class on CIFAR-10. Boldface indicates the backdoor target class (class 0). MAD = Median Absolute Deviation used for outlier detection.}
\label{tab:nc_norms}
\begin{tabular}{lcccccc}
\toprule
\textbf{Quant.} & \textbf{Class 0} & \textbf{Class 1} & \textbf{Class 2} & $\cdots$ & \textbf{Median} & \textbf{MAD} \\
\midrule
FP32 & \textbf{215.6} & 296.5 & 275.9 & $\cdots$ & 282.1 & 15.4 \\
INT8 & 27.9 & 27.9 & 27.9 & $\cdots$ & 27.9 & 0.0 \\
INT4 & 222.9 & 284.9 & 189.5 & $\cdots$ & 246.1 & 33.9 \\
\bottomrule
\end{tabular}
\end{table}

\subsection{Hyperparameters}

\textbf{Model Training:} SGD optimizer with learning rate 0.1, momentum 0.9, weight decay $5\times10^{-4}$, batch size 128. CIFAR-10 trains for 20 epochs; GTSRB for 15 epochs.

\textbf{BadNet Attack:} $3\times3$ white square trigger positioned at bottom-right corner, 10\% poisoning rate, target class 0.

\textbf{Neural Cleanse:} 1000 optimization steps per class, learning rate 0.1, $\ell_1$ regularization coefficient $\lambda=0.01$, MAD outlier threshold 2.0.

\textbf{Activation Clustering:} $k$-means with $k=2$ clusters, silhouette score threshold 0.15 for anomaly detection.

\textbf{STRIP:} 100 perturbation samples per input, entropy threshold at 5th percentile of clean distribution, model flagged if $>5\%$ samples fall below threshold.

\textbf{Spectral Signatures:} Top eigenvalue ratio threshold for outlier detection.

\textbf{Fine-Pruning:} Dormancy threshold 0.1\% of neurons.

\end{document}